\documentclass[sigconf]{acmart}

\def\BibTeX{{\rm B\kern-.05em{\sc i\kern-.025em b}\kern-.08emT\kern-.1667em\lower.7ex\hbox{E}\kern-.125emX}}

\copyrightyear{2019}
\acmYear{2019}
\setcopyright{acmlicensed}
\acmConference[KDD '19]{25th ACM SIGKDD Conference on Knowledge Discovery and Data Mining}{August 04--08, 2019}{Anchorage, Alaska}
\acmPrice{}
\acmISBN{}
\acmDOI{}

\usepackage[english]{babel}
\addto\extrasenglish{%
}

\usepackage{wrapfig}

\begin{document}
	
	%
	% The "title" command has an optional parameter, allowing the author to define a "short title" to be used in page headers.
	\title{The Error is the Feature: How to Forecast Lightning using a Model Prediction Error}
	
	\subtitle{[Applied Data Science Track, Category Evidential]}
	
	%
	% The "author" command and its associated commands are used to define the authors and their affiliations.
	% Of note is the shared affiliation of the first two authors, and the "authornote" and "authornotemark" commands
	% used to denote shared contribution to the research.

	\author{Christian Sch{\"o}n}
	\affiliation{%
		\institution{Saarland Informatics Campus}
		\department{Big Data Analytics Group}
	}
	
	\author{Jens Dittrich}
	\affiliation{%
		\institution{Saarland Informatics Campus}
		\department{Big Data Analytics Group}
	}

	\author{Richard M{\"u}ller}
	\affiliation{%
		\institution{German Meteorological Service}
		\city{Offenbach}
		\country{Germany}
	}

\begin{abstract}
	Despite the progress within the last decades, weather forecasting is still a challenging and computationally expensive task. Current satellite-based approaches to predict thunderstorms are usually based on the analysis of the observed brightness temperatures in different spectral channels and emit a warning if a critical threshold is reached.
	Recent progress in data science however demonstrates that machine learning can be successfully applied to many research fields in science, especially in areas dealing with large datasets. We therefore present a new approach to the problem of predicting thunderstorms based on machine learning.
	The core idea of our work is to use the error of two-dimensional optical flow algorithms applied to images of meteorological satellites as a feature for machine learning models. We interpret that optical flow error as an indication of convection potentially leading to thunderstorms and lightning.
	To factor in spatial proximity we use various manual convolution steps. We also consider effects such as the time of day or the geographic location. We train different tree classifier models as well as a neural network to predict lightning within the next few hours (called \textit{nowcasting} in meteorology) based on these features. In our evaluation section we compare the predictive power of the different models and the impact of different features on the classification result. 
	Our results show a high accuracy of 96\% for predictions over the next 15 minutes which slightly decreases with increasing forecast period but still remains above 83\% for forecasts of up to five hours. The high false positive rate of nearly 6\% however needs further investigation to allow for an operational use of our approach.
\end{abstract}

	

	%
	% Keywords. The author(s) should pick words that accurately describe the work being
	% presented. Separate the keywords with commas.
	
	%
	% This command processes the author and affiliation and title information and builds
	% the first part of the formatted document.
	\maketitle
	
	\section{Introduction}
	\label{sec:Introduction}
	
	Weather forecasting is a very complex and challenging task requiring extremely complex models running on large supercomputers. Besides delivering forecasts for variables such as the temperature, one key task for meteorological services is the detection and prediction of severe weather conditions. Thunderstorms are one such phenomenon often accompanied by heavy rain fall, hail, and strong wind. However, predicting them and giving precise information about their severity and moving direction is a hard task.
	
	Current state-of-the-art systems such as \textit{NowCastMIX}~\cite{james2018nowcastmix}, a system operated by the Deutscher Wetterdienst, combine a multitude of data sources to generate warnings with a spatial resolution of 1$\times$1 km and a temporal resolution of five minutes for severe weather conditions such as thunderstorms. Brightness temperatures of spectral satellite channels and their differences exceeding a certain threshold are interpreted as a sign for critical conditions potentially leading to thunderstorms. Radar systems are used to detect water particles in the atmosphere and clouds potentially developing towards thunderstorms whereas numerical weather prediction (NWP) models offer estimations of the near storm environment. Lightning detection systems allow us to localize thunderstorms by measuring electric waves. Even using such advanced models, the prediction of thunderstorms remains very challenging, especially for forecast periods larger than one hour when the false alarm ratio increases to more than 80\%.	The key to thunderstorm forecasting is the early and precise detection of convection.
	
%	Current state-of-the-art systems operated by weather services to forecast thunderstorms include information based on brightness temperatures of spectral satellite channels and their difference, where values exceeding a certain threshold are interpreted as a sign for critical conditions potentially leading to thunderstorms.
%	Radar systems are used to detect water particles in the atmosphere and clouds potentially developing towards thunderstorms whereas numerical weather prediction (NWP) models offer estimations of the near storm environment. Lightning detection systems allow us to localize thunderstorms by measuring electric waves. State-of-the-art systems such as \textit{NowCastMIX}~\cite{james2018nowcastmix}, a system operated by the Deutscher Wetterdienst, include all these measurements to generate warnings with a spatial resolution of 1$\times$1 km and a temporal resolution of five minutes for severe weather conditions such as thunderstorms. Even using such advanced models, the prediction of thunderstorms remains very challenging, especially for forecast periods larger than one hour when the false alarm ratio increases to more than 80\%.	The key to thunderstorm forecasting is the early and precise detection of convection.
	
	Satellite data, which is nowadays part of many weather forecasting products and offered a significant performance boost for larger forecast periods, however is not yet established for the operational prediction of developing thunderstorms  even though it offers a combination of high spatial and temporal resolution, two key elements for a successful forecast. Our approach therefore investigates a new way to predict thunderstorm clouds based on satellite data. The core idea of our work is to use the prediction error of a first prediction model as a feature. That error-feature is then used for the actual, second, different prediction model. In more detail, we compute the error of two-dimensional optical flow algorithms applied to images of meteorological satellites and interpret it as an indication of possible movement in the third dimension (convection) potentially leading to thunderstorms and lightning.
	
	Our main contributions to the problem of forecasting thunderstorms are the following:
	\begin{enumerate}
	\item We present a prediction model to forecast lightning using satellite images. The core idea of our work is to use the prediction error of a first prediction model as a feature for a second (different) prediction model. Although learning from the error of a prediction model reminds of boosting, our approach differs from such methods as described in \autoref{ssec:CoreIdea}.
	
	\item	We present a new set of features based on the error of optical flow algorithms applied to satellite images in different channels which can be used to predict thunderstorms. These features differ from previously used approaches mainly based on temperature differences as presented in \autoref{sec:Thunderstorms}. Based on this feature set, we then apply different machine learning algorithms in order to automatically predict the occurrence of lightning. This process is described in more detail in \autoref{sec:Approach}.
		
%	\item	Considering the immediate future, i.e.~the next 15 minutes, our model achieves accuracy values of more than 96\%. Especially features based on convolution with large kernel sizes have shown a positive impact on the predictive power of the models. Even for larger forecast periods of up to five hours, the accuracy remains above 83\% showing promising results for applications in nowcasting. A detailed description of the results can be found in \autoref{sec:Results}.
	
	\item	An evaluation of these error-based features used to predict the immediate future, i.e.~the next 15 minutes, show the importance of convolution, especially with large kernel sizes. This evaluation also motivates the use of a broad range of channels. A more detailed description of the results can be found in \autoref{sec:Results}.

%	\item We present experiments showing the capability of tree classifier to predict lightning based on features derived from the error of optical flow algorithms on a balanced test set containing lightning with a forecast period of 15 minutes.  add additional features such as time of day and coordinates to further improve the accuracy. An evaluation of larger forecast periods of up to five hours finally shows the limits of the approach, however still reaching impressive accuracy values.
	
	\item We present experiments showing the capability of machine learning models to predict lightning based on features derived from the error of optical flow algorithms as well as additional, satellite-based features on a balanced test set containing lightning with a forecast period of 15 minutes up to 5 hours. The results show an accuracy of more than 96\% for the best model on the smallest period, decreasing to 83\% for the largest period.
	\end{enumerate}

	This paper is structured as follows: \autoref{sec:Thunderstorms} gives a quick recap of the physics of thunderstorms followed by a review of current forecasting systems in \autoref{sec:RelatedWork}. We introduce our new approach in \autoref{sec:Approach}, followed by the necessary data preparation and feature generation steps presented in \autoref{sec:Pipeline}. \autoref{sec:ExperimentalSetup} briefly introduces the experimental setup before we present the first set of experiments and results in \autoref{sec:Results} based on the approach described before. \autoref{sec:AddingFeatures} introduces additional features beyond pure error-based attributes and evaluates the impact of this enhancement compared to the original approach. In \autoref{sec:Increasing}, we present the performance of our approach for larger forecast periods.
	
	\section{A Quick Recap of Thunderstorms}
	\label{sec:Thunderstorms}
		
	In this section, we briefly review the physics of thunderstorms. Thunderstorms belong to a class of weather phenomena called \textit{convective systems}, characterized by an updraft of warm air from lower levels of the atmosphere towards higher and colder levels, accompanied by a downdraft allowing the cool air to flow back towards the ground. Thunderstorms essentially form a special case of convective systems which are characterized by a strong updraft of warm, moist air which freezes in the upper atmosphere leading to electric load inequalities and hence lightning.
	
	A basic understanding of convection and the emergence of thunderstorms is crucial to identify potential features that can be used in machine learning algorithms. Section~\ref{ssec:Emergence} therefore gives a short introduction to the basic effects associated with thunderstorms. 
		
	\subsection{The Three Phases of a Thunderstorm}
	\label{ssec:Emergence}
		
	The emergence of thunderstorms is usually divided into three separate phases: the developing stage, the mature stage, and finally the dissipating stage~\cite{bott2016synoptische,mogil2007extreme}.
	
\noindent\textbf{Developing Stage.}
	In the first stage of a thunderstorm emergence, warm and moist air is rising to upper levels of the atmosphere. This so called updraft can be caused by geographic reasons such as air rising at the slopes of mountains, but also by thermal reasons such as large temperature differences between the ground and the upper atmosphere, especially in summer. Given that the convection is strong enough, the warm and moist air will eventually pass the dew point where it starts to condensate, forming a so called \textit{cumulus congestus} which is characterized by strong updrafts offering a fresh supply of warm, moist air. During this first phase, the condensed air might form first precipitation particles which however do not yet fall to the ground.
	
\noindent\textbf{Mature Stage.}
	In the second stage of the thunderstorm, air cools down in the troposphere, one of the upper layers of the atmosphere. If the system is strong enough, the moist air will eventually pass the point where it starts to freeze and sinks back to the ground at the sides of the updraft. This leads to a so called \textit{downdraft}, a strong wind formed by falling (and potentially evaporating) precipitation. The \textit{horizontal divergence} at the top of the cloud leads to the typical anvil form called \textit{cumulonimbus}. This process of rising and finally freezing moist air is also responsible for the emergence of lightning. Cold, frozen particles may split into smaller, negatively charged particles and larger ones with a positive charge. The smaller the particle, the faster it will rise in the updraft and finally fall down back to the ground. This potentially leads to a separation of negative charge in the downdraft and positive charge in the updraft, which results in an electric load imbalance, which in turn may result in lightning which decreases this imbalance.
	
\noindent\textbf{Dissipating Stage.}
	In the last stage of a thunderstorm, the updrafts loose strength and finally stop. Due to the missing supply of warm, moist air, the system becomes unstable and eventually breaks down. Precipitation looses intensity, mainly consisting of the remaining condensed air at the top of the cloud.
	
\section{Related Work}
\label{sec:RelatedWork}
	
Satellite-based thunderstorm forecast is usually based on the analysis of the observed atmospheric brightness temperature in different spectral channels. If the brightness temperatures or brightness temperature differences reach a critical threshold, the forecast system emits a thunderstorm warning. The satellite-based methods are often supported by NWP stability indices, which are used as an indicator for the potential energy of a system.
%Thunderstorm forecasts are often based on indices describing the state of the atmosphere and its tendency to develop into a thunderstorm. The detection of strong convection eventually leading to thunderstorms is mainly based on 
%temperature differences (estimated using radio sounding or NWP models potentially guided by satellite images) which are used as an indicator for the potential energy of a system. 
Common indices used are the \textit{Convective Available Potential Energy (CAPE)}~\cite{moncrieff1976dynamics}, the \textit{Lifted Index (LI)}~\cite{galway1956lifted} or the \textit{KO Index}~\cite{koindex}.
All these indices essentially consider the potential temperature at different levels of the atmosphere (described by pressure). The greater the differences, the more likely the atmosphere will become unstable and develop a convective system potentially leading to a thunderstorm.

Besides using such index-based forecasting methods, state-of-the-art systems such as NowCastMIX include additional data sources: Radar systems are used to detect precipitation and frozen water particles in the atmosphere. Specialized systems such as \textit{KONRAD}~\cite{lang2001cell} and \textit{CellMOS}~\cite{hoffmann2008entwicklung} try to detect thunderstorm cells and predict their movement. Lightning detection systems such as \textit{LINET}~\cite{Betz2009Linet} measure electric waves in the atmosphere to identify lightning and their location. NWP models such as \textit{COSMO-DE}~\cite{baldauf2018beschreibung} are used to identify potentially interesting areas in advance and to model the local near storm environment. A combination of these systems is then used to predict storms and follow their movement.
	
Although machine learning algorithms are not yet widely used in meteorology, there are first attempts to apply them to weather prediction and partially also to thunderstorm forecasting. Ruiz and Villa tried to distinguish convective and non-convective systems based on logistic regression and Random Forest models~\cite{ruiz2008storms}. 
The dataset consisted of different features derived from satellite images, in particular exact values and differences for temperatures and gradients. Similar approaches have been presented by Williams et al., trying to predict convective initiation with a forecast period of one hour~\cite{williams2008combining,williams2008machine} and Ahijevych et al.~\cite{williams2016probabilistic} for a forecast period of two hours. They trained Random Forests with datasets consisting of different kinds of features, covering raw satellite image values, derived fields such as CAPE or LI and radar data. Veillette et al.~\cite{veillette2013convective} used features based on satellite images, NWP models and environmental information to train various machine learning models on the prediction of convective initiation.
	
Although these approaches use similar machine learning algorithms, they differ from our approach in the feature set used for training. To the best of our knowledge, there has been no approach so far which is based on using the error of nowcasting algorithms for satellite images such as optical flow to predict thunderstorms.
	
\section{The Error is the Feature}
\label{sec:Approach}

In this section we introduce our approach to predict lighting using a model prediction error as the feature. We describe the core idea as well as the meteorological background. 

\subsection{Core Idea}
\label{ssec:CoreIdea}

\begin{figure}
		\centering
		\includegraphics[trim=5mm 5mm 22mm 5mm, keepaspectratio, clip, width=0.865\linewidth]{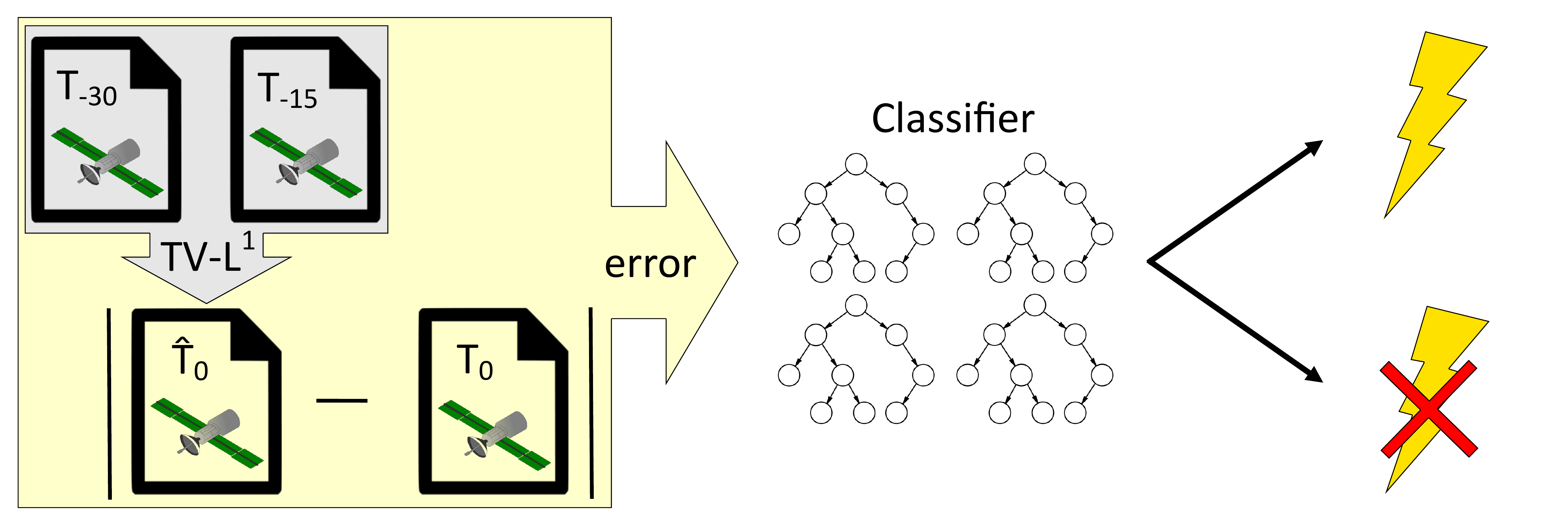}
		\caption[Overview Approach]{Overview of our approach: Two consecutive satellite images at $T_{-30}$ and $T_{-15}$ are read to predict the next image $\hat{T_0}$ using TV-L$^1$. The error is then computed as the absolute difference between $\hat{T_0}$ and the original image $T_0$. 
%		Then we compute the error by taking the absolute difference between $\hat{T_0}$ and the original image $T_0$. 
		We use that error to predict lightning based on different classifier.}
		\label{fig:OverviewApproach}
\end{figure}

Rather than directly deriving features from satellite images, our approach is based on the error resulting from forecasting the next image based on previous images using optical flow algorithms such as \textit{TV-L$^1$}~\cite{zach2007duality}.
A high-level overview of our approach is shown in \autoref{fig:OverviewApproach}.
The core idea can be formulated as follows: The movement of air within the atmosphere is a \textit{three}-dimensional phenomenon with the third dimension depicted as brightness on satellite images. Optical flow algorithms predicting future images based on past observations however can only detect and predict \textit{two}-dimensional movements. The error resulting from the application of optical flow might therefore be related to the vertical movement of clouds, a sign for convection potentially leading to thunderstorms. Machine Learning algorithms can then be used to learn the relation between these error values and the occurrence of lightning which is in turn a sign of thunderstorms.

Although learning from the error of a model might remind of boosting in the first place, our approach differs from such methods: Boosting algorithms iteratively train a weak classifier on a given data set and then apply a re-weighting of the samples in order to minimize the prediction error of the next weak classifier trained on the same data set. Our approach in contrast does not try to improve the optical flow model used to predict satellite images as it would be done in boosting. Instead, we interpret the error of this first model as a feature to train a second, different model to predict lightning. The two models differ significantly in their input and output. Although both methods, boosting and our approach, learn from the error of a prediction model, we would not call our approach a boosting method, but a two stage approach instead.
	
\subsection{Meteorological Background}

The SEVIRI instrument~\cite{schmetz2002introduction} on-board of the second generation of Meteosat satellites (MSG) generates multichannel images of Europe, Africa, and the surrounding seas every 15 minutes with a sub-satellite resolution of 3$\times$3 km for most channels.
The different channels of the SEVIRI instrument are designed to measure different wave lengths. 
The first three channels as well as the high resolution channel essentially measure the reflection of solar light at clouds or the Earth's surface, which is closely related to the \textit{albedo} of the repective objects.
%The first three channels as well as the high resolution channel essentially measure the \textit{albedo}, i.e. the reflection of solar light at clouds or the earth's surface. 
The higher the intensity of the reflected light, the brighter the tile appears on the image. As clouds have a high albedo, they appear bright in these channels. The remaining channels of the SEVIRI instrument differ from these first channels mainly in the fact that they measure primarily infrared, i.e. thermal radiation. According to Kirchhoff's law of thermal radiation, the emissivity is equal to the absorptivity for every body which absorbs and emits thermal radiation~\cite{kirchhoff1860ueber}. Planck's law postulates that the intensity of the emission depends on the temperature of the body: The warmer the body, the higher the intensity~\cite{planck1900zur}. This relation can be used to estimate the (brightness) temperature of the surface of objects, e.g. of cloud tops, allowing us to observe changes in the (brightness) temperature using infrared satellite channels. 
%Changes in the (brightness) temperature of cloud tops can therefore be observed with satellite channels in the infrared range of light. 
A central feature of cloud convection is the vertical updraft of moist air which leads to an increase of the cloud albedo and cooling of the cloud, thus, to changes in the intensities. However, convective clouds move also horizontally which makes it difficult to relate changes of the observed intensities to the vertical updraft: Changes could result from real vertical updraft or from artefacts induced by differences in intensities between the Earth's surface and clouds. 
The central problem is the lack of three-dimensional cloud movement data as SEVIRI provides only two-dimensional images.
%The central problem is that information about the cloud movement in three-dimensions is needed for the direct detection of convection. SEVIRI however provides only two-dimensional images. 
Thus the vertical updraft can not be simply derived by the calculation of the difference in intensities between subsequent satellite images in time, preventing the detection of convection.
%In contrast to the first group of channels based on the albedo, the remaining channels use an inverse mapping of intensity to brightness: The lower the intensity of the thermal radiation, the brighter the tile. We can use this knowledge to determine the approximate temperature of surfaces using satellite imaginary. Changes in the brightness of tiles on satellite images in these channels therefore directly reflect changes in the temperature of the reflecting surface.

Optical flow algorithms such as TV-L$^1$ however assume that the intensity of objects does not change between consecutive images. Errors resulting from the application of optical flow on satellite images can therefore result from two reasons: The first reason is an inaccurate forecast of the movement of clouds, meaning that the actual position of a cloud object on an image was different from the predicted position. This error essentially results from weaknesses of the optical flow algorithm. Errors might however also result from cloud objects whose position was predicted correctly, but whose brightness changed between two consecutive images. As explained above, these brightness changes directly relate to the temperature of the clouds which in turn relates to their height as clouds become colder the higher they are.
These (rapidly) rising clouds are however exactly the result of convection leading to thunderstorms. Detecting them would allow us to predict thunderstorms in advance, earlier than lightning detection systems or radar systems searching for precipitation particles.

\section{Data Preprocessing Pipeline}
	\label{sec:Pipeline}

As machine learning models rely on carefully engineered features in order to learn relations to some target variables, we present in this section the steps which are necessary to transform the raw data to the features used in our models.

\subsection{Basic Feature Generation \& Target Values}
\label{ssec:Preprocessing}
	
The raw input used for our approach essentially consists of two data sources: binary files containing MSG images for the different channels and CSV files containing lightning detected by the \textit{LINET} system. As \textit{LINET} only covers Europe, satellite data representing surrounding areas can be ignored. The first step in our pipeline was therefore a projection of the satellite images to the area covered by \textit{LINET} using the \textit{Satpy} library~\cite{Satpy}. We decided to use the first nine channels of the satellite, covering a spectrum from the visible light at $0.6 \mu m$ wavelength up to the infra red light at $10.8 \mu m$~\cite{schmetz2002introduction}.
	
	In the second step, the error values are computed in the following way: Two consecutive satellite images at $T_{-30}$ and $T_{-15}$ minutes were fed into the \textit{TV-L$^1$} algorithm to compute the next image of this sequence $\hat{T_0}$. The implementation was based on the \textit{OpenCV} python library~\cite{OpenCV}, the parameters used are given in \autoref{tab:Parameters_TVL1} in the appendix. 
%	As the satellite images are available every 15 minutes, we essentially take the images at $T_{-30}$ and $T_{-15}$ minutes to generate the image $\hat{T_0}$. 
	In addition, we read the original image at $T_0$ and compute the absolute difference between these two images which we consider the error of our nowcasting. Using a classifier, we compute a target variable $\hat{Y}_{15}$ predicting  the occurence of lightning within the next 15 minutes.
	\begin{equation*}
	error_0 =| T_0~~ -~~ \hat{T}_0 | = |  T_0~~ - \text{\textit{TV-L$^1$}}(T_{-30}, T_{-15}) |.
	\end{equation*}
	\begin{equation*}\label{lightningclassifier}
		\hat{Y}_{15} = \text{classifier}(error_0).
	\end{equation*}

The lightning data obtained from the \textit{LINET} network was supplied as CSV files containing as attributes the time lightning occurred as well as its location, charge, and height. We transformed it to geographically tiled (or rastered) maps where each tile stores the number of lightning occurring at this location in time interval $[T; T_{+15})$ (``within the next 15 minutes''). 
Each tile now represents data with a temporal extension of 15 minutes and a spatial resolution equal to the one of MSG.
%The maximum geographic resolution achievable using the Meteosat satellite is 3$\times$3 km for the majority of the channels, however due to the earth's curvature, the resolution of a tile in Europe is a little worse.
%The temporal extension of a tile was 15 minutes, i.e.~$\hat{Y}_{15}$ indicates whether lightning occurs within time interval $[T; T_{+15})$ (``within the next 15 minutes'').

For performance reasons, the images containing the error values as well as the maps containing the lightning were stored as binary numpy arrays to avoid the repetition of the costly load/parse and spatial projection operations before each training step.
	
	\begin{figure}
		\centering
		\includegraphics[trim=34mm 37mm 260mm 10mm, clip, width=0.435\linewidth]{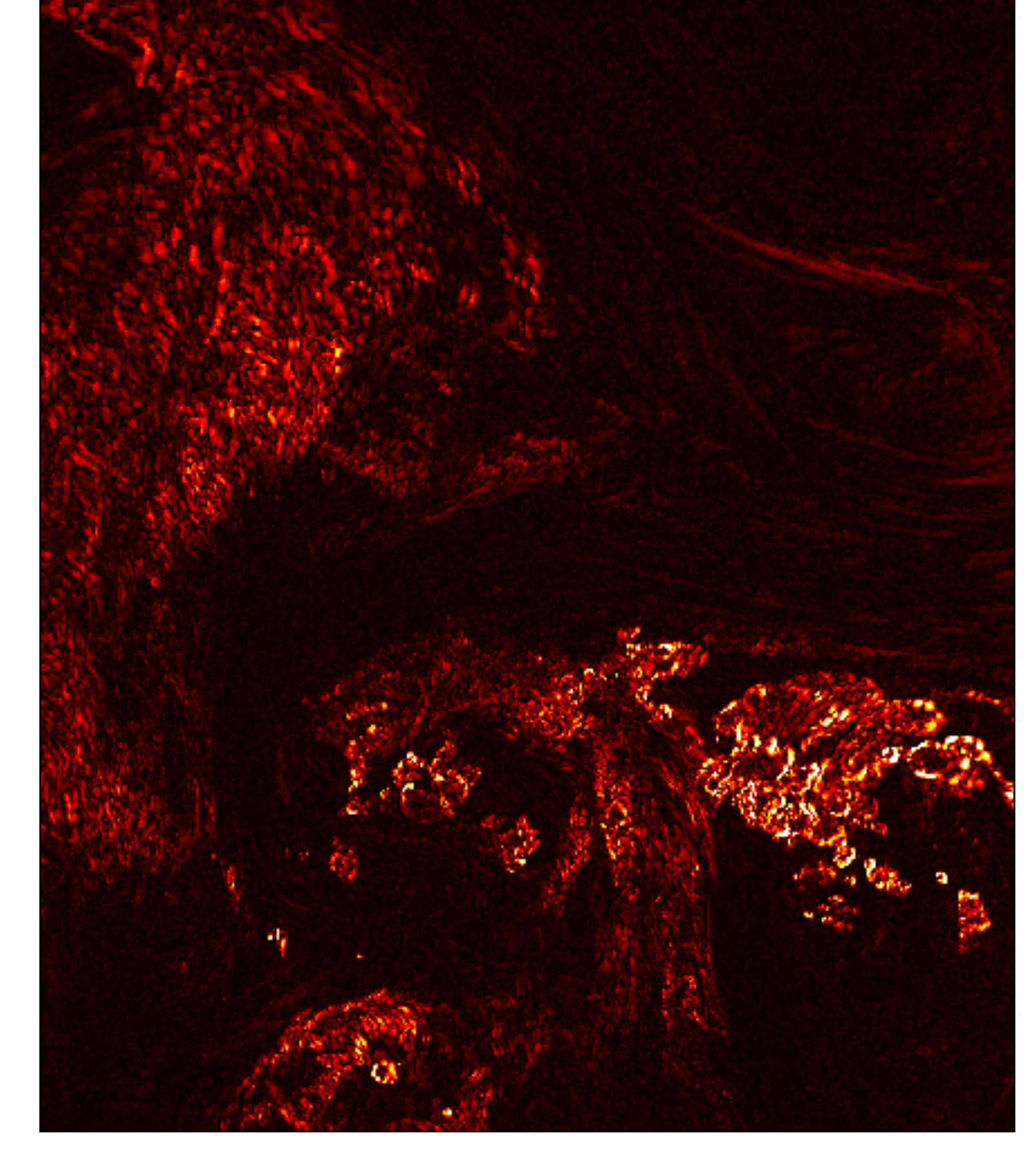}
		\includegraphics[trim=260mm 37mm 34mm 10mm, clip, width=0.435\linewidth]{WV06X_2017_06_01_13_15_quantile_0_9995.pdf}
		\caption[Images After Preprocessing]{Example images after preprocessing: The error in channel WV6.2 at 2017-06-01 13:15 as heatmap \& the lightning data for the following 15 minutes as binary mask.}
		\label{fig:Images After Preprocessing}
		\vspace*{-.4cm}
	\end{figure}

\subsection{Enhancing Features Using Convolution}
\label{ssec:Feature}
	
Based on our general assumption that high error values indicate the presence of lightning, we use the absolute error values in each of the first nine channels of SEVIRI as our basic features. \autoref{fig:Images After Preprocessing} shows an optical comparison between the error in channel WV6.2 and the occurence of lightning, supporting this assumption.
	
As clouds and thunderstorms move over time, we also include features covering the spatial influence from nearby tiles. 
The use of convolution on the images containing the error values allows us to identify thunderstorms entering a tile from a neighbouring tile.
%This allows us to identify thunderstorms entering a tile from a neighbouring tile. This information was created using convolution on the images containing the error values. 
We decided to explore kernels of size 3$\times$3 up to 9$\times$9 to cover local lightning events as well as larger areas.
	
For each kernel size, we explored four different convolution operations: we decided to use the maximum and minimum operations over the window specified by the kernel size to identify areas with extreme values. In addition, we explored the average value over different kernel sizes to cover information about the overall state within a certain area. Finally, we added a convolution based on the gaussian distribution to weight error values near the centre of the kernel higher than values at the border, i.e.~this corresponds to a weighted average. The intuition behind this approach is that geographically close locations should influence the result more than locations at larger distances. The weights are given by the following formula where x and y denote the coordinates with respect to the centre of the kernel and $\sigma$ the standard deviation:

\begin{equation*}
	G(x,y,\sigma) = \frac{1}{2\pi\sigma^2} e^{-\frac{x^2+y^2}{2\sigma^2}}.
\end{equation*}

Our implementation was based on the following filters defined in the \textit{ndimage} package of the \textit{SciPy} Python library~\cite{Scipy-Ndimage-Filter}: \textit{maximum\_filter}, \textit{minimum\_filter}, \textit{uniform\_filter} and a combination of \textit{gaussian\_kernel} and \textit{convolve} for the last operation.	

This combination of four convolution kernel sizes \textit{times} four convolution operations \textit{times} nine channels each \textit{plus} the raw error values of each channel yields a total of $4 * 4 * 9 + 9 = 153$ features.

We decided to train the classifiers on a per tile basis, i.e. the models were not trained on two-dimensional images, but on single tiles and their corresponding feature values. The target values were binary values indicating the presence or absence of lightning at this tile given a specific offset compared to the time stamp of the error values.

\section{Experimental Setup}
\label{sec:ExperimentalSetup}

The experiments conducted in this thesis were based on 4-fold cross-validation, which allowed us to combine three sets for training while using the remaining set for testing. Our dataset approximately covers one month, from 2018-06-01 00:00 to 2018-07-04 06:30. We decided to split this set along the time axis, using a twelve hour margin between the sets. As training with highly imbalanced data is very hard, we decided to use downsampling on a per image basis, meaning that we took all tiles with a lightning event present on an image and chose the same number of tiles without lightning event at random from the same image. Due to various data errors during the extraction of the raw, binary satellite data and the preprocessing steps, we could not consider all images. The resulting cross-validation sets and the sizes of the classes are shown in \autoref{tab:Cross_Validation_Sets_Lightning_Forecast}. A more detailed explanation of this setup and the versions of the different Python libraries used can be found in \autoref{sec:Experimental_Setup_Details}.
	
	\begin{table}
		\caption[Cross-Validation Sets Lightning Forecast]{Main characteristics of the cross-validation sets.}
		\label{tab:Cross_Validation_Sets_Lightning_Forecast}
		\begin{tabular}{ccc}
			\toprule
			\textbf{\#} & \textbf{Time Range} & \textbf{\# samples /  class} \\
			\midrule
			0 & 2017-06-01 00:30 to 2017-06-08 23:00 & 624,877 \\
			1 & 2017-06-09 11:00 to 2017-06-17 09:45 & 65,261 \\
			2 & 2017-06-17 21:45 to 2017-06-25 20:15 & 256,456 \\
			3 & 2017-06-26 08:15 to 2017-07-04 06:30 & 291,956 \\
			\bottomrule
		\end{tabular}
	\end{table}

\section{Results}
\label{sec:Results}
	
The main assumption of our approach is the correlation between high error values and the existence of lightning. A simple statistical evaluation gives evidence for this assumption: \autoref{tab:Comparison_error_distribution_lightning} shows the distribution of the error values in channel WV6.2 for the timespan 2018-06-01 00:30 to 2018-06-03 06:15. The first column refers to all tiles on all images within this time range while the following two columns represent the two classes, namely the \textit{no-lightning} and the \textit{lightning} class with respect to the next 15 minutes. Comparing these values, we can state that the mean error for the \textit{lightning} class is higher than the one for the \textit{no-lightning} class. Simply assuming that a high error value necessarily indicates the presence of lightning is, however, wrong as the overall maximum belongs to the \textit{no-lightning} class. Nevertheless, we can still conclude that high values more likely indicate the presence of lightning than the absence.
	
	\autoref{tab:Comparison_error_distribution_lightning} also shows the extreme imbalance of the dataset. Only 169,912  samples out of more than 230 million belong to the \textit{lightning} class, resulting in a fraction of only 0.074\% which decreases to 0.066\% if computed over all available data.
	
	\begin{table}
		\caption[Comparison of the Error Distribution for tiles With and Without Lightning]{Nowcasting error distribution (2018-06-01 00:30 to 2018-06-03 06:15) in channel WV6.2, per class distribution taken with respect to lightning within the next 15 minutes.}
		\label{tab:Comparison_error_distribution_lightning}
		\begin{tabular}{cccc}
			\toprule
			& \textbf{All tiles} & \textbf{No-Lightning} & \textbf{Lightning} \\
			\midrule
			\textbf{\# samples} & 230,036,544 & 229,866,632 & 169,912 \\
			\textbf{mean error} & 0.5141 & 0.5128 & 2.3600 \\
			\textbf{std. deviation} & 0.7680 & 0.7627 & 2.8515 \\
			\textbf{min error} & 0.0 & 0.0 & 0.0 \\
%			\textbf{25\% quantile} & 0.1385 & 0.1385 & 0.5419 \\
%			\textbf{50\% quantile} & 0.2922 & 0.2922 & 1.3701 \\
%			\textbf{75\% quantile} & 0.5885 & 0.5872 & 3.0280 \\
			\textbf{max error} & 30.8376 & 30.8376 & 30.0355 \\
			\bottomrule
		\end{tabular}
	\end{table}

	\subsection{Predicting the Immediate Future}
	\label{ssec:Immediate}
	
	To check whether our assumption is correct and we can indeed train machine learning models to predict lightning, we first conducted an experiment to forecast the immediate future: Given the error resulting from the satellite images at $T_{-30}$, $T_{-15}$ and $T_0$, can we predict whether there will be lightning within the next 15 minutes, i.e.~between $T_0$ and $T_{+15}$?
	
	We decided to use tree-based classifiers for this first set of experiments as they provide more transparency in understanding how a prediction was made compared to neural networks. The models chosen for evaluation are simple Decision Trees as well as ensemble methods based on them: Random Forests as described by Breiman~\cite{breiman2001random}, Gradient Boosting as introduced by Mason et al.~\cite{mason2000boosting} and AdaBoost as presented by Freund and Schapire~\cite{freund1997decision}. The implementations are based on the \textit{Scikit-Learn} Python library~\cite{scikit-learn}: \textit{DecisionTreeClassifier}, \textit{RandomForestClassifier}, \textit{AdaBoostClassifier} and~\textit{GradientBoostingClassifier}. For most parameters we used their default values as specified in the documentation. We adopted some parameters to avoid overfitting and limit the training time. The most important parameters are depicted in~\autoref{app:TreeClassifier}.
	
\autoref{tab:Accuracy} shows the resulting accuracy values for the different models and test sets. As expected, the simple Decision Tree shows the worst performance in each test set, reaching values between 84\% and 89\%. Gradient Boosting  performed best with accuracy results between 89\% and 92\%. The last row of the table shows the accuracy over all test sets, underlining a strong Gradient Boosting model, a weak Decision Tree model and the Random Forest as well as AdaBoost models in between.	
Considering the computational effort required to create these models, notice however that the improvement of about 2\% between Random Forests and Gradient Boosting comes at the cost of a much higher training time: For the largest training set consisting of the cross-validation sets 0, 2, and 3, the training phase for  Random Forest which could be conducted in parallel on all CPU cores took about 14 minutes. Gradient Boosting however which can by nature not run in parallel took about 818 minutes which is more than 54 times the duration of Random Forest.
	
	\begin{table}
		\caption[Accuracy for Lightning Prediction]{The accuracy for each model and test set: Decision Tree (DT), Random Forest (RF), AdaBoost (AB), Gradient Boosting (GB). The last row shows the accuracy over all sets.}
		\label{tab:Accuracy}
		\begin{tabular}{ccccc}
			\toprule
			\textbf{Test Set} & \textbf{DT} & \textbf{RF} & \textbf{AB} & \textbf{GB} \\
			\midrule
			\textbf{0} & 87.494\% & 87.661\% & 90.203\% & 90.973\% \\
			\textbf{1} & 86.549\% & 87.219\% & 88.648\% & 89.090\% \\
			\textbf{2} & 89.946\% & 91.724\% & 90.744\% & 92.714\% \\
			\textbf{3} & 84.118\% & 90.050\% & 85.483\% & 90.500\% \\
			\textbf{overall} & 87.156\% & 89.042\% & 89.120\% & 91.123\% \\
			\bottomrule
		\end{tabular}
	\end{table}
	
As a high overall accuracy does however not necessarily indicate an equally good performance on both classes.,
we also present the main characteristics resulting from the confusion matrices for all models in \autoref{tab:Comparison_Confusion_Matrices}, created by summing up the results on the individual test sets. 
Especially Gradient Boosting performs equally well on both classes, leading to high precision and recall values while minimizing the false alarm rate. Random Forest  does even slightly outperform Gradient Boosting  in terms of recall, but at the cost of a much worse performance in terms of precision. All models except AdaBoost tend to achieve better results for the \textit{lightning} class, which seems slightly easier to learn. 
	
	\subsection{Limitations}
	\label{ssec:Limitations}
	
%	Although these results seem a promising start, the current models will likely not be suitable for operational thunderstorm forecasting. Especially the high false positive rate which still is in the range of 10\% for the best model needs further investigation. 
	These results indicate the potential of the method for operational usage, however, the false positive rate of 10\% for the best model needs further investigation and improvement.
	In addition, one should keep in mind that the results shown here are based on a balanced subset of all tiles. Using the original, highly unbalanced data would not affect the values for accuracy, recall and false positive rate, but lead to a much lower precision due to the larger amount of false positives (given the assumption that the subsample is representative for the real data). The total amount of samples per class over all test sets is 1,238,550. Considering the fact that only roughly 0.066\% of all tiles belong to the Lightning class and we chose the same amount of samples from the No-Lightning class, we can conclude that the overall amount of tiles in the images considered for this paper having a negative true condition is roughly in the region of 1,876,590,909. Considering the False Positive Rate of 10.62\% for the Gradient Boosting model, this calculation would lead to a precision of roughly 0.57\% if all tiles were considered. Applying this model alone in practice to issue warnings would therefore not be feasible. We can however calculate the false positive rate FPR necessary to achieve a better precision PR the following way:
	
	\begin{equation*}
		PR = \frac{TP}{TP + FP} = \frac{TP}{TP + FPR * N} \Leftrightarrow FPR = \frac{(1-PR) * TP}{PR * N}.
	\end{equation*}
	
	In the formula above, TP denotes the number of true positives and N the overall number of tiles with a negative true condition.
%	as it has the same number of samples than the lightning class. 
	However, even assuming a perfect recall, a model which aims at a precision of 20\% (the value currently achieved by NowCastMIX for a one hour offset) would have to achieve a false positive rate of roughly 0.26\% which seems extremely hard to achieve.

	\begin{table}
		\caption{Main characteristics of the models over all test sets: precision (PR), recall (RE) and false positive rate (FPR).}
		\label{tab:Comparison_Confusion_Matrices}
		\begin{tabular}{ccccc}
			\toprule
			& \textbf{DT} & \textbf{RF} & \textbf{AB} & \textbf{GB} \\
			\midrule
%			\textbf{TP} & 1,095,563 & 1,160,234 & 1,097,896 & 1,150,155 \\
%			\textbf{FP} & 175,166 & 193,120 & 128,844 & 131,507 \\
%			\textbf{TN} & 1,063,384 & 1,045,430 & 1,109,706 & 1,107,043 \\
%			\textbf{FN} & 142,987 & 78,316 & 140,654 & 88,395 \\
			\textbf{PR} & 86.22\% & 85.73\% & 89.50\% & 89.74\% \\
			\textbf{RE} & 88.46\% & 93.68\% & 88.64\% & 92.86\% \\
			\textbf{FPR} & 14.14\% & 15.59\% & 10.40\% & 10.62\% \\
			\bottomrule
		\end{tabular}
	\end{table}

	\subsection{Feature Evaluation}
	\label{ssec:FeatureEvaluation}
	
	The models considered so far all belong to the class of tree classifier algorithms which essentially all work in a very similar way: In each step, the model chooses a feature and a corresponding threshold and performs a binary split of the set of samples into two distinct subsets assigned to the left and right subtree. The decision which feature and threshold to choose is taken according to the gini impurity which measures the impurity of a node where lower values indicate purer nodes. Given the relative frequencies $p_i$ of each class $i \in \{1,...,J\}$, the gini impurity is defined as:
	
	\begin{equation*}
	g(p) = 1 - \sum_{i=1}^{J} p_i^2.
	\end{equation*}
	
	Based on this impurity measure, we can define a feature importance measure. Given the weights of the individual trees $\alpha_t$ in the ensemble, the total number of samples $N$, the number of samples $N_k, N_L, N_R$ reaching a parent node $k$ as well as its left and right child node and the feature $s(k)$ used in the split of node $k$, we define the gini importance $FI(x_i)$ of some feature $x_i$ as:
	
	\begin{equation*}
		FI(x_i) = \sum_{t} \alpha_t \sum_{k, s(k) = x_i} \frac{N_k}{N} \left( g\left(p^k\right) - \frac{N_L}{N_k} g\left(p^L\right) - \frac{N_R}{N_k} g\left(p^R\right) \right).
	\end{equation*}
	
\begin{figure}
		\centering
		\includegraphics[trim=2.5mm 3.0mm 3.7mm 2.6mm, clip, width=0.950\linewidth, keepaspectratio]{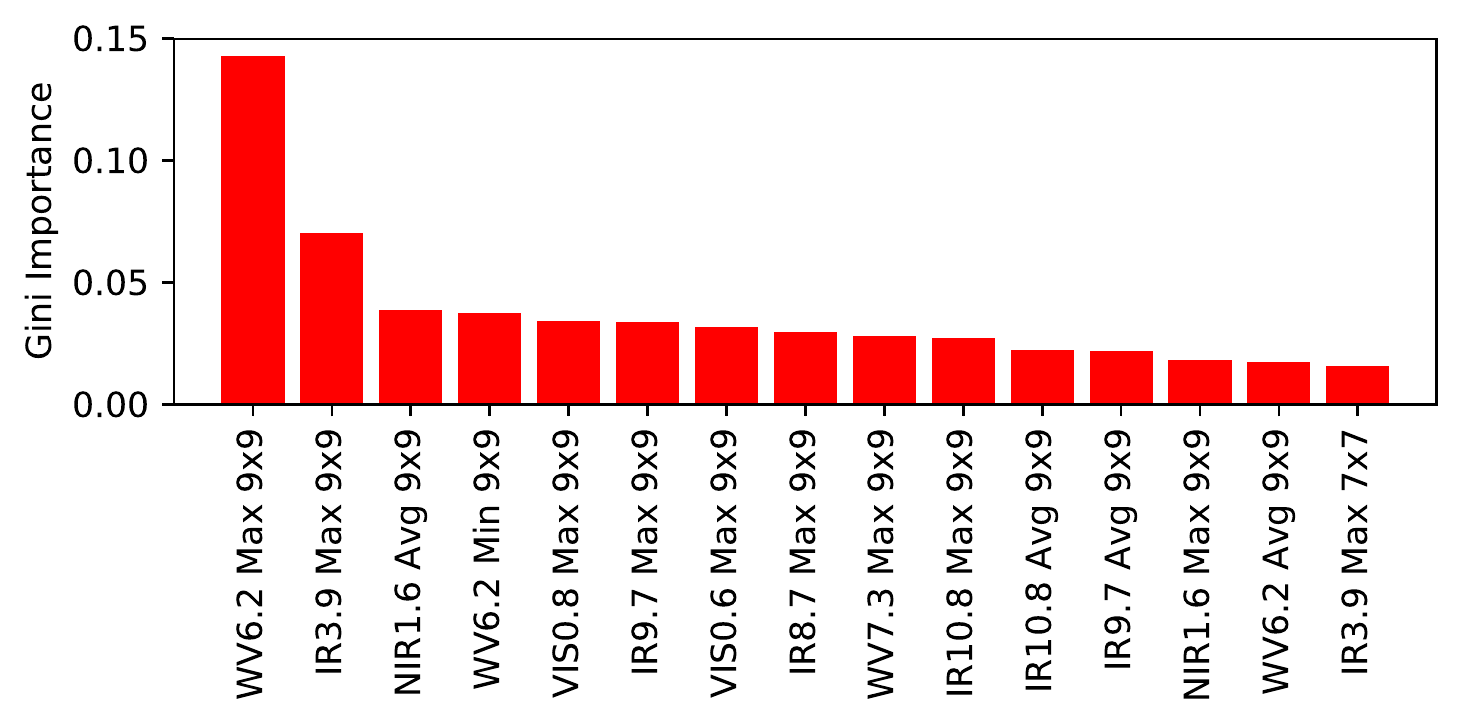}
		\caption[Feature Importance Gradient Boosting Top 15 Training Sets 0, 2, 3]{The top 15 features and their gini importance for the Gradient Boosting model trained on the sets 0, 2 and 3.}
		\label{fig:Feature Importance Gradient Boosting Top 35}
	\end{figure}

\autoref{fig:Feature Importance Gradient Boosting Top 35} shows an example plot for the top 15 features of Gradient Boosting  trained on the sets 0, 2, and 3 which together form the largest training set. 
The maximum value within a 9$\times$9 tile window of channel WV6.2 is the most important feature with a gini importance of about 0.14, followed by the maximum convolution with the largest kernel in channel IR3.9, however reaching just half the importance of the first feature.
%According to this evaluation, the maximum value within a 9$\times$9 tile window of channel WV6.2 is the most important feature, reaching a total gini importance of about 0.14. The second ranked feature uses a maximum convolution with the largest kernel in channel IR3.9, however reaching just half the importance of the first feature. 
The third and following features keep loosing importance with values between 0.04 and 0.01. The feature importances for the other three training sets show very similar results: 
The maximum values of a 9$\times$9 kernel in channels WV6.2 and IR3.9 remain the prominent features, with the first ranked feature reaching twice the gini importance of the second ranked one, followed by slightly decreasing importances for the following features. The results of Random Forest and AdaBoost models look very similar.
%The maximum values for a 9$\times$9 kernel in the channels WV6.2 and IR3.9 remain the prominent features ranking either first or second. The feature on the first rank roughly reaches twice the gini importance of the second ranked one, followed by slightly decreasing importances for the following features. The results of Random Forest and AdaBoost models look very similar.
Decision Trees rely on a much smaller subset of features: Compared to Gradient Boosting, the top ranked feature shows twice the gini importance with values of 0.35 to 0.6. 
%Decision Trees rely on a much smaller subset of features: The top ranked feature has a gini importance of 0.35 to 0.6, more than twice the value of the top ranked feature for Gradient Boosting. 
%The features on the following ranks show a fast drop of the importance values, indicating that the model weights them much lower considering their discriminative power. Models based on an ensemble of Decision Trees such as Random Forests, AdaBoost, and Gradient Boosting obviously favour a broad range of comparably strong features over a small subset of extremely discriminative features which potentially explains their improved accuracy.
The features on the following ranks show a fast drop of the importance values, indicating that the model relies on a small subset of features whereas ensemble methods such as Random Forests, AdaBoost, and Gradient Boosting obviously favour a broader range of features which potentially explains their improved accuracy.
	
\autoref{tab:featureUsage} shows the number of occurrences for channels, kernel sizes as well as convolution types within the top 35 features of the Gradient Boosting model for the different training sets. Considering the channels, we can clearly state that IR3.9 belongs to the most prominent channels, appearing six to eight times among the top features, followed by IR9.7 and VIS0.8. There is however no clear concentration on a specific subset of channels: Usually all channels appear in all training sets, except for one set where IR8.7 does not appear within the top 35 features. 
Considering the different kernel sizes, the model clearly favours larger kernels over smaller ones. The 1$\times$1 column which means taking the exact error value of a single tile shows no usage at all. The largest kernel size of 9$\times$9 however is always used at least 20 times, i.e.~in nearly two third of all features. 
A similar observation holds true for the convolution types where the maximum convolution clearly dominates with 15 to 21 appearances within the top features. In contrast, the gaussian convolution does not appear at all, indicating that this convolution type does not help the model to discriminate the samples.
	
	\begin{table}
		\caption{Channel, kernel size and convolution type usage within the top 35 features of Gradient Boosting depending on the training sets.}
		\label{tab:featureUsage}
		\begin{tabular}{ccccc}
			\toprule
			\textbf{Feature} & \textbf{sets 1,2,3} & \textbf{sets 0,2,3} & \textbf{sets 0,1,3} & \textbf{sets 0,1,2} \\
			\midrule
			\textbf{VIS0.6} & 4 & 4 & 4 & 4 \\
			\textbf{VIS0.8} & 5 & 5 & 5 & 3 \\
			\textbf{NIR1.6} & 2 & 2 & 2 & 2 \\
			\textbf{IR3.9} & 6 & 8 & 8 & 8 \\
			\textbf{WV6.2} & 4 & 4 & 4 & 4 \\
			\textbf{WV7.3} & 2 & 2 & 4 & 3 \\
			\textbf{IR8.7} & 3 & 2 & 0 & 1 \\
			\textbf{IR9.7} & 5 & 5 & 5 & 7 \\
			\textbf{IR10.8} & 4 & 3 & 3 & 3 \\
			\midrule
			\textbf{1$\times$1} & 0 & 0 & 0 & 0 \\
			\textbf{3$\times$3} & 0 & 2 & 3 & 2 \\
			\textbf{5$\times$5} & 2 & 3 & 4 & 4 \\
			\textbf{7$\times$7} & 8 & 7 & 8 & 9 \\
			\textbf{9$\times$9} & 25 & 23 & 20 & 20 \\
			\midrule
			\textbf{Max} & 15 & 18 & 21 & 20 \\
			\textbf{Min} & 12 & 9 & 6 & 8 \\
			\textbf{Avg} & 8 & 8 & 8 & 7 \\
			\textbf{Gaussian} & 0 & 0 & 0 & 0 \\
			\bottomrule
		\end{tabular}
	\end{table}

\section{Adding non-error-based Features}
\label{sec:AddingFeatures}
	
The experiments in Section~\ref{sec:Results} were based only on features derived from the \textit{error} of the optical flow algorithm. What happens if we add other features? That is what this section is about.
	
\noindent\textbf{Raw Image Values.} Error values do not allow for any inference on the values of the original satellite image, making it hard to distinguish cloudy and non-cloudy tiles with similar error values. We therefore added the original values of the different channels as additional features which indeed allow for such a separation. We again use the assumption that convolution incorporating information of surrounding tiles might be helpful and therefore apply the same convolutions as for the error values of the optical flow.
	
\begin{figure}
	\includegraphics[trim=4.0mm 4.8mm 5.1mm 9.5mm, clip, width=\linewidth, keepaspectratio]{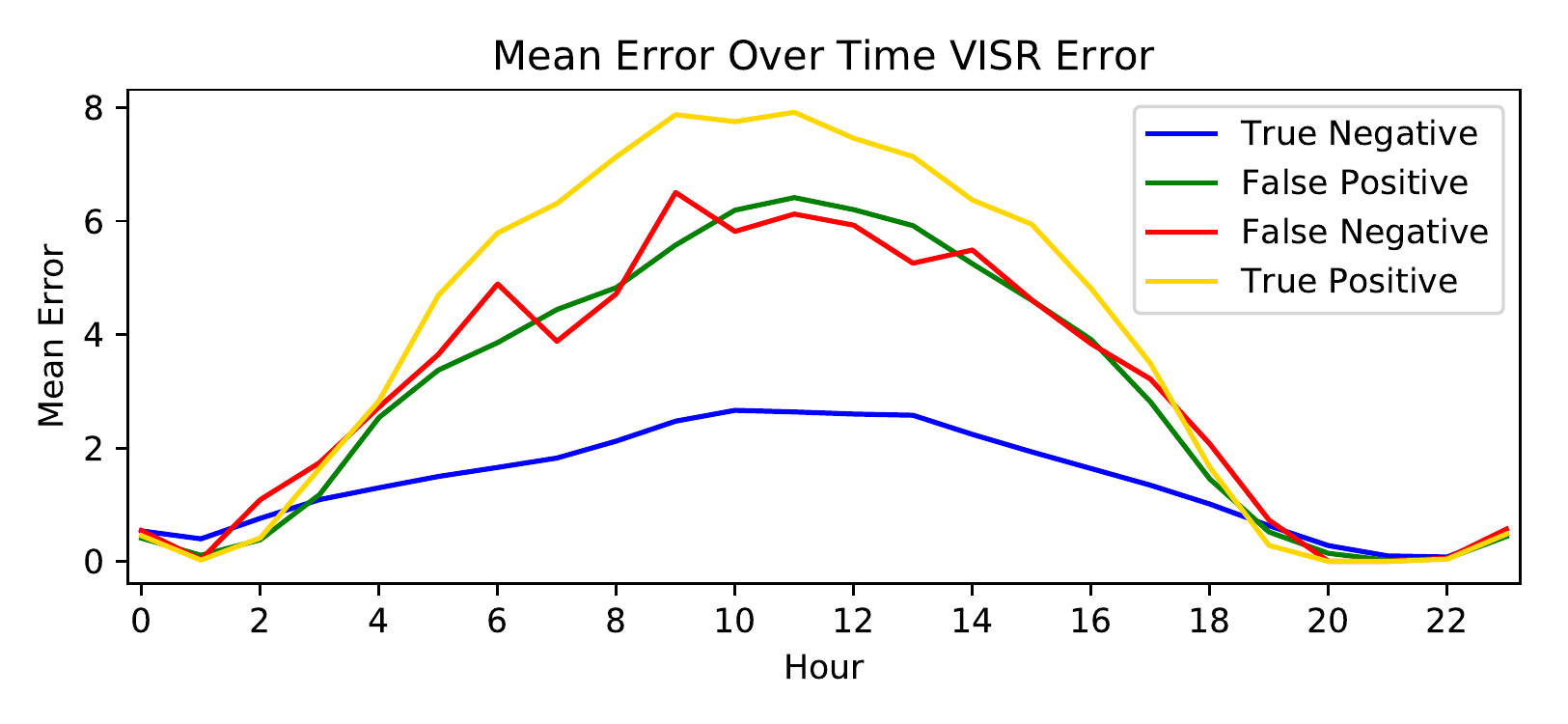}
	\caption{Mean error values for the VIS0.6 channel over time based on the results of Gradient Boosting.}
	\label{fig:VIS0.6_Values_Over_Time}
\end{figure}

\noindent\textbf{Time Of Day.} The first three channels of SEVIRI cover the range of the visible light. These channels essentially show a black image over night, giving no useful information for thunderstorms occuring in the late evening or early morning. \autoref{fig:VIS0.6_Values_Over_Time} shows the mean values for the VIS0.6 channel over time clearly indicating very low values over night giving no useful information to distinguish the two classes. At noon however the values of this channel allow for a pretty clear distinction of lightning and no-lightning tiles. The distribution of lightning over time shows that the average number of lightning per 15 minutes stays below 400 during night and early morning before it starts to rise at approximately 10:00, reaching a peak at 15:00 with more than 1700 lightning per 15 minutes.
Leaving out these channels would therefore eliminate a very useful data source for the distinction in the afternoon when most lightning occur. We therefore added an additional feature depicting the time of day as four digit number in the format [hhmm] to allow the model to take decisions based on time.

\begingroup
\setlength{\columnsep}{11pt}%
\setlength{\intextsep}{3pt}%
	\begin{wrapfigure}{r}{0.350\linewidth}
		\includegraphics[trim=10mm 58mm 806mm 31mm, clip, width=\linewidth, keepaspectratio]{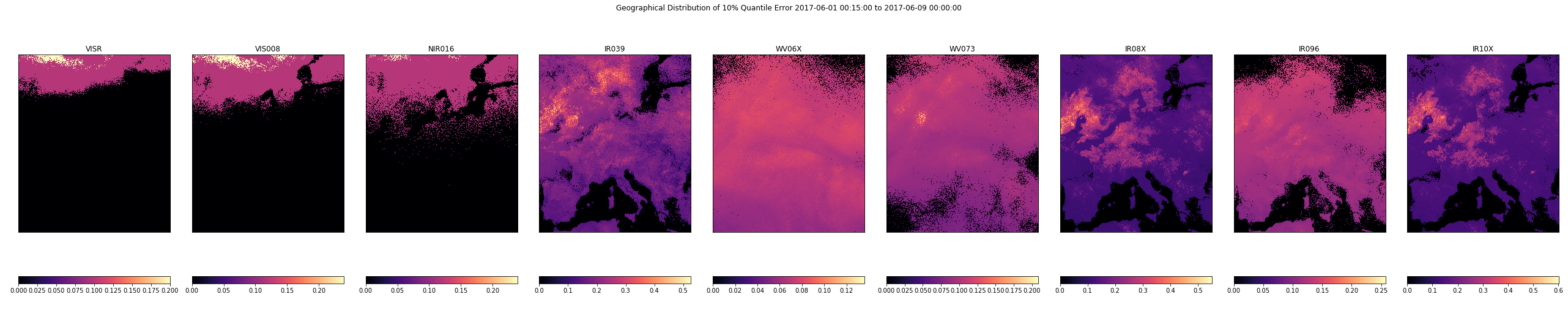}
		\caption{Geographical distribution of the 10\% quantile error (channel VIS0.6).}
		\label{fig:VIS0.6_10_quantile_error}
	\end{wrapfigure}
\noindent\textbf{Coordinates.} 
During sunrise, the channels covering the visible light will become brighter from north to south as the sunlight reaches tiles near the pole earlier than tiles near the equator, given the same longitude. During sunset, the same phenomenon occurs in the oppsite direction as the tiles become darker from south to north. 
\autoref{fig:VIS0.6_10_quantile_error} shows the geographical distribution of the 10\% quantile error clearly indicating that tiles in the lower half of the image remain black for a longer period of time. 
We therefore also include the x and y coordinate of a tile on the image as additional features to allow for a distinction of such cases during sunset or sunrise. Adding such geographic information might also help to distinguish mountains and plains, potentially having different triggers for thunderstorms.

\endgroup

\noindent\textbf{Removing Features.} To avoid an exploding number of features, we eliminated the gaussian convolution as it showed no improvement in the previous experiments. Considering the kernel sizes, we decided to use only two kernels, namely a 7$\times$7 and a 13$\times$13 kernel to incorporate the result that the previous models preferred larger kernels over smaller ones. 
We still keep the values for the single tile to allow the models to learn from differences between the value of the tile itself and convolution values which include surrounding tiles.
This leads to a total of 129 features: The error values resulting from nowcasting as well as the raw satellite image values, each combined with 3 convolution types and two kernel sizes plus the features for the time of day and the coordinates.
	
	The new feature set was fed to a Random Forest model with slightly adapted parameters compared to the first experiment: We conducted a hyperparameter search with \textit{GridSearchCV} from the Scikit-Learn library on the largest training set using three folds. We varied the number of estimators in the interval [50, 300], the maximum depth in the interval of [14, 26] and the minimum number of samples per leaf in the interval [3, 81]. Out of the 64 configurations tested, a configuration with 300 estimators, a maximum depth of 26 and a minimum number of samples per leaf of three yields the best results on this largest training set. However, this configuration tends to start overfitting the data on smaller trainings sets. We therefore decided to use a configuration with the maximum depth set to 18, the minimum number of samples per leaf set to nine and the number of estimators set to 200. This configuration has shown a slightly worse performance (96.00 \% accuracy compared to 96.24\%), but does not tend to overfit too fast on smaller training sets.
	
\begin{figure}
		\includegraphics[trim=3.9mm 9.1mm 5.1mm 5.1mm, clip, width=0.930\linewidth, keepaspectratio]{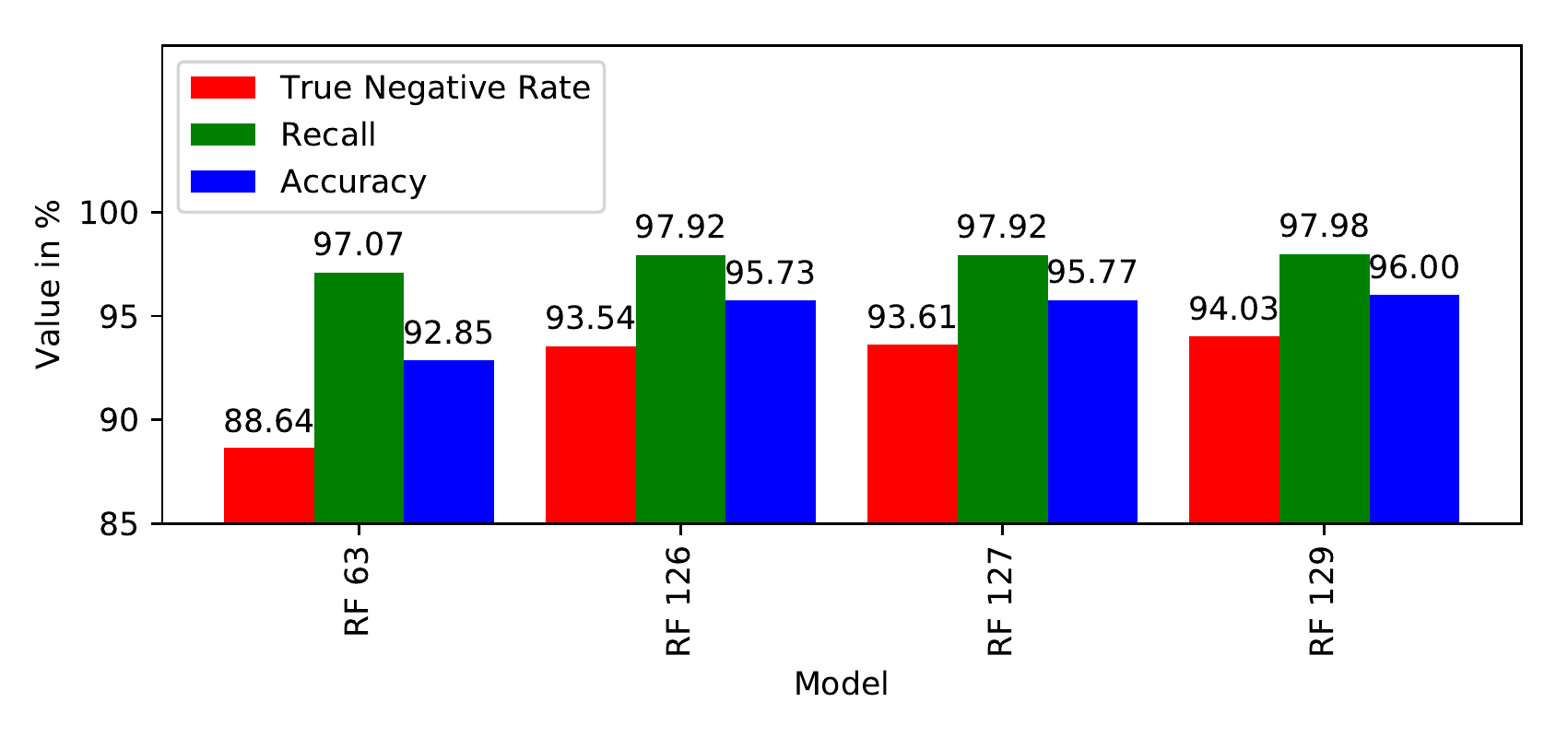}
		\caption[Accuracy, Recall, TPR Additional Features]{Main metrics for Random Forests using only error-based features (RF63), adding raw satellite images (RF126), time of day (RF127) and coordinates (RF129).}
		\label{fig:Accuracy_Additional_Features}
	\end{figure}

	To evaluate the impact of each of the new features, we conducted several experiments, adding the new features one after the other. \autoref{fig:Accuracy_Additional_Features} shows the resulting values for accuracy, recall and true negative rate.
	 RF63 denotes a Random Forest just using the 63 features based on the error of the optical flow as baseline. RF126 is a Random Forest including the values for the raw satellite images, resulting in 126 features. RF127 then adds the time of day while RF129 finally uses all features including the coordinates.
	
	This new setup increased the accuracy up to 96.00\%, clearly outperforming the Gradient Boosting model reaching only 91.12\%. Considering the per class performance, this new model reaches a recall of 97.98\% and a false positive rate of 5.97\%, showing an equal boost for both classes compared to the previous results.
	
	\section{Increasing the Forecast Period}
	\label{sec:Increasing}
	
	Forecasting lightning just for the next 15 minutes is probably the easiest task, but not necessarily the most useful as it gives little to no time to react. We therefore decided to conduct additional experiments to evaluate if our approach does also offer the possibility to increase the forecast period up to five hours.
	The basic setup remains the same, especially the data preprocessing and feature generation steps. Instead of taking the maps which include the lightning for the next 15 minutes, we now used a specific offset to determine the maps used as target values for our machine learning models. An offset of +0:00:00 essentially just means the same as in the previous sections, namely considering the lightning for the next 15 minutes. An offset of +5:00:00 however now indicates that the model is trained to learn the presence or absence of lightning five hours in the future, i.e.~between +5:00:00 and +5:15:00 compared to the last available satellite image.
	
	\autoref{tab:Lightning Forecast Accuracy Random Forest Depending On Offset} shows the results of the corresponding experiments for different models. RF1 is the Random Forest model from the first set of experiments, using the 153 original features only considering the error values of nowcasting. RF2 is the Random Forest model described in the previous section, i.e. using 129 features.
	
	\begin{table}
		\caption[Lightning Forecast Accuracy Depending on Offset]{Comparison of the forecast accuracy for different models and offsets from 0 to 5 hours on all test sets.}
		\label{tab:Lightning Forecast Accuracy Random Forest Depending On Offset}
		\begin{tabular}{cccc}
			\toprule
			\textbf{Offset} & \textbf{RF1} & \textbf{RF2} & \textbf{NN1} \\
			\midrule
			\textbf{+0:00:00} & 89.04\% & 96.00\% & 96.42\% \\
			\textbf{+1:00:00} & 84.61\% & 92.25\% & 92.19\% \\ 
			\textbf{+2:00:00} & 80.29\% & 86.30\% & 88.03\% \\
			\textbf{+3:00:00} & 76.67\% & 81.35\% & 85.48\% \\
			\textbf{+4:00:00} & 73.68\% & 80.82\% & 84.43\% \\
			\textbf{+5:00:00} & 71.39\% & 78.36\% & 83.87\% \\
			\bottomrule
		\end{tabular}
	\end{table}

All the previous sections only used tree classifiers as prediction models. We decided to also test a neural network for comparison. NN1 is a neural network based on the Keras framework~\cite{Keras} using 9 hidden, dense layers with sizes [64, 64, 64, 32, 32, 32, 16, 16, 16] and the leaky ReLU activation function. The single output node uses the sigmoid activation. The model was trained using the Adam optimizer and binary cross-entropy as loss function. The training time was limited to 500 epochs with early stopping if the accuracy on the test set did not improve by at least 0.01\% over the last 50 epochs. The batch size was fixed to 25,000 and the feature set identical to the one used for RF2, except for the fact that all features were normalized to the unit interval $[0;1]$. We decided to not consider convolutional networks as the highly imbalanced data would most likely lead to models just predicting black images indicating no lightning at all, reaching an accuracy of more than 99.9\% this way.
	
	\begingroup
	\setlength{\columnsep}{10pt}%
	\setlength{\intextsep}{2pt}%
	\begin{wrapfigure}{r}{0.52\linewidth}
%	\begin{figure}
		\includegraphics[trim=3.9mm 4.8mm 4.1mm 3.9mm, clip, keepaspectratio, width=\linewidth]{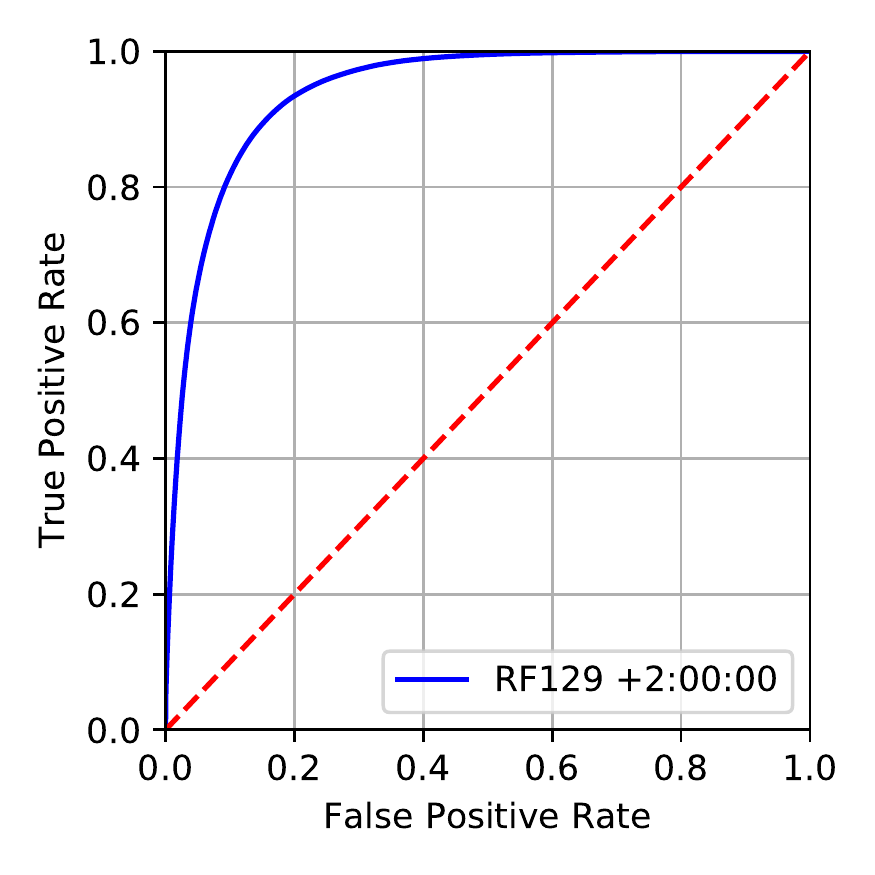}
		\caption{ROC curve for the model RF2 with offset +2:00:00.}
		\label{fig:ROC_RandomForest}
%	\end{figure}
	\end{wrapfigure}
Considering the performance, we must state the the models lose accuracy with increasing offset. However, this result is explainable by the fact that the weather is a very chaotic phenomenon where accurate forecasts become more and more difficult with increasing forecast period. But even with the largest offset tested, the results clearly remain above the 50\% margin one would expect from random guessing on a balanced dataset. An accuracy of still more than 83\% shows promising results for future work. However, a larger offset comes at the cost of an increasing number of false positives further reducing the precision of the model. This problem still needs to be solved. 
The receiver operating characteristic (ROC) curve of model RF2 for a two hour offset depicted in \autoref{fig:ROC_RandomForest} however shows the capability of our model to learn the problem, reaching an area under the curve (AUC) value of 0.94. This is clearly better than the results reported by Ahijevych et al.~\cite{williams2016probabilistic} and Veillette et al.~\cite{veillette2013convective}. One should however keep in mind that their models were trained on predicting newly initiating convective systems which slightly differs from our approach covering all appearances of lightning.
The increasing gap between the models RF2 and NN1 most likely relates to overfitting in RF2 as it is weaker on smaller training sets, but stronger on larger ones. With more data available, we expect the Random Forest model to become competitive to the neural network.
	
	\endgroup

	\section{Conclusion}
	\label{sec:Conclusion}
	
	The results of our approach seem very promising. Using just error values resulting from the nowcast of satellite images based on optical flow, different tree classifiers could be trained to predict lightning in the immediate future with an accuracy of more than 91\%. 
	Features using convolution with larger kernel sizes show the greatest impact on the accuracy of the models. The results of our models are based on a broad range of channels which are all present within the top 35 features. 
	Adding non-error-based features improved the accuracy further, now reaching values of up to 96\%. Even for the largest forecast period of five hours considered in this paper, the accuracy of the models still remain above 83\% which is clearly better than the 50\% which one would expect from random guessing. Comparing the results to those of Ahijevych et al. ~\cite{williams2016probabilistic} or Veillette et al.~\cite{veillette2013convective}, our approach clearly demonstrates lower false positive rates given the same recall, leading to a superior overall performance. However, the high false positive rate still needs further investigation to prepare the approach for potential future operational use.
	
	Neural Networks as presented in the last section could offer an alternative to the tree-based models mostly used in this paper. Considering the data used as basis, we currently limit our approach to features directly related to satellite images. Including other data sources as it is done in NowCastMIX will most likely increase the performance and reduce the high false positive rate. Especially our boosted models might also benefit from a dimensionality reduction of the feature set by elimination of less important features which would allow us to grow deeper models in reasonable time, potentially yielding better results.
	
	%
	% The acknowledgments section is defined using the "acks" environment (and NOT an unnumbered section). This ensures
	% the proper identification of the section in the article metadata, and the consistent spelling of the heading.
	\begin{acks}
		We would like to thank Deutscher Wetterdienst (DWD) and nowcast GmbH for providing us the necessary data for our research. We would also like to thank Stéphane Haussler from DWD for providing us the Python scripts to read the binary satellite data and Jilles Vreeken for giving us helpful feedback.
	\end{acks}
	
	%
	% The next two lines define the bibliography style to be used, and the bibliography file.
	\bibliographystyle{ACM-Reference-Format}
	\bibliography{bibliography}
	
	\newpage
	% 
	% If your work has an appendix, this is the place to put it.
	\appendix
	
	\section{Experimental Setup Details}
	\label{sec:Experimental_Setup_Details}
	
	All Experiments were conducted on a computer equipped with two Intel Xeon E5-2600 v4 processors, 32 GB of RAM and two Nvidia GTX 1080 Ti graphics cards. The implementation was based on Python 2.7.14\footnote{As the OpenCV implementation provided by DWD was based on Python 2.} using Satpy 0.9.2 to read and project the raw satellite data. The optical flow computation was based on OpenCV 3.4.0.12 and the models on Scikit-Learn 0.19.1.
	
	Our training was based on a 4-fold cross-validation, meaning that the complete set of available data points was divided into four sets (folds) where three sets were combined to form the training data and the remaining set was used to test the performance of the trained model. Splitting could have been done along the geographic coordinates (meaning that we reserve a specific part of each image for testing) or along the time axis (meaning that we reserve images at a specific timestamp for testing). Splitting along the geographic coordinates would however carry the risk of overlapping training and testing sets due to the convolution operations applied during feature generation. We therefore decided to use the second approach, namely a splitting along the time axis. Our dataset approximately covers one month, from 2018-06-01 00:30 to 2018-07-04 06:30. Simply splitting this time range into training and test set at some point in time called $T_S$ would lead to the problem that data points taken from the last image before $T_S$ and data points taken from the first image after $T_S$ could be highly correlated as the weather conditions did not change much during these 15 minutes. To avoid such problems, the cross-validation sets were designed such that the start and end points of the different sets are separated by a twelve hour margin. Taking into account all available tiles would have led to highly imbalanced sets as more than 99.9\% of all tiles belong to the \textit{no-lightning} class (compare \autoref{tab:Comparison_error_distribution_lightning}). Models trained with such imbalanced data would tend to always predict the absence of lightning, achieving a trivial (and useless) accuracy of more than 99.9\% this way. To avoid this problem, we decided to balance the training and test sets in the following way: Balancing through downsampling was done on a per image basis, meaning that we took all tiles with a lightning event present on an image and chose the same number of tiles without lightning event at random from the same image.
	
	Due to various data errors during the extraction of the raw, binary satellite data and the preprocessing steps, we could not consider all images. The resulting cross-validation sets and the sizes of the classes are shown in \autoref{tab:Cross_Validation_Sets_Lightning_Forecast}. 
	
	\section{Model Parameters}
	
	For the sake of reproducibility , we denote in this section the most relevant parameters of the different models especially if they differ from the default settings described in the documentation of the frameworks used. All parameters not mentioned in this section where left on their default values.
	
	\subsection{TV-L$^{1}$}
	\label{app:TVL1}
	
	The implementation of TV-L$^1$ used in this paper was based on the DualTVL1OpticalFlow class contained in the OpenCV library. The parameters used are given in \autoref{tab:Parameters_TVL1}. We have not conducted a hyperparameter search for the optical flow ourselves, but used parameters proposed by the German Meteorological Service which tested several configurations.
	
	\begin{table}
		\caption[Parameters TV-L$^1$]{Parameters for the TV-L$^1$ algorithm used for optical flow computation.}
		\label{tab:Parameters_TVL1}
		\begin{tabular}{cc}
			\toprule
			\textbf{Parameter} & \textbf{Value} \\
			\midrule
			\textbf{tau} & 0.1 \\
			\textbf{lambda} & 0.0005 \\
			\textbf{theta} & 0.3 \\
			\textbf{epsilon} & 0.001 \\
			\textbf{outerIterations} & 10 \\
			\textbf{innerIterations} & 30 \\
			\textbf{gamma} & 0.0 \\
			\textbf{nscales} & 7 \\
			\textbf{scaleStep} & 0.5 \\
			\textbf{warps} & 5 \\
			\textbf{medianFiltering} & 1 \\
			\bottomrule
		\end{tabular}
	\end{table}
	
	\subsection{Tree Classifier}
	\label{app:TreeClassifier}
	
	The tree classifier implementations were all based on the scikit-learn library using the classes \textit{DecisionTreeClassifier}, \textit{RandomForestClassifier}, \textit{AdaBoostClassifier} and \textit{GradientBoostingClassifier}. The most important parameters used for the first set of experiments with 153 features based on the error of the optical flow are depicted in \autoref{tab:Parameters_Tree_Classifier}. Parameters not mentioned here were left on their default values. A dash indicates that this parameter is not present for this model. For the Random Forest models described in later sections of this paper, some parameters have been adapted: The max\_depth parameter was increased to 18 and the minimum samples per leaf was fixed to an absolute value of nine instead of using a percentage of the training set size.
	
	\begin{table}
		\caption[Parameters Tree Classifier]{Parameters of the different tree classifiers. A dash means this parameter is not present for the specific model.}
		\label{tab:Parameters_Tree_Classifier}
		\begin{tabular}{ccccc}
			\toprule
			\textbf{Parameter} & \textbf{DT} & \textbf{RF} & \textbf{AB} & \textbf{GB} \\
			\midrule
			\textbf{Max depth} & 12 & 12 & 6 & 7 \\
			\textbf{Criterion} & gini & gini & gini & gini \\
			\textbf{Min \# samples / leaf} & 0.01\% & 0.01\% & 0.01\% & 0.01\% \\
			\textbf{\# estimators} & - & 200 & 50 & 100 \\
			\textbf{\# jobs} & - & 16 & - & - \\
			\textbf{Learning rate} & - & - & 1.0 & 0.1 \\
			\textbf{Boosting criterion} & - & - & - & Friedman MSE \\
			\bottomrule
		\end{tabular}
	\end{table}
	
\end{document}